\documentclass{bmvc2k}

\pdfoutput=1
\usepackage{multirow}
\usepackage[inline]{enumitem}
\usepackage{booktabs}
\usepackage{amsmath}
\usepackage{amssymb}
\usepackage{ctable}

\title{3D-LMNet: Latent Embedding Matching for Accurate and Diverse 3D Point Cloud Reconstruction from a Single Image}

\addauthor{Priyanka Mandikal\textsuperscript{*}}{priyanka.mandikal@gmail.com}{0}
\addauthor{K L Navaneet\textsuperscript{*}}{navaneetl@iisc.ac.in}{0}
\addauthor{Mayank Agarwal\textsuperscript{*}}{mayankgrwl97@gmail.com}{0}
\addauthor{R. Venkatesh Babu}{venky@iisc.ac.in}{0}

\addinstitution{
Video Analytics Lab,\\
Department of Computational and\\ Data Sciences,\\
Indian Institute of Science,\\
Bangalore, India
}

\runninghead{Mandikal, Navaneet, Agarwal, Babu}{3D Point Cloud Reconstruction}

\def\etal{\emph{et al}\bmvaOneDot}

\begin{document}

\maketitle

\begin{abstract}

3D reconstruction from single view images is an ill-posed problem. Inferring the hidden regions from self-occluded images is both challenging and ambiguous. We propose a two-pronged approach to address these issues. To better incorporate the data prior and generate meaningful reconstructions, we propose 3D-LMNet, a latent embedding matching approach for 3D reconstruction. We first train a 3D point cloud auto-encoder and then learn a mapping from the 2D image to the corresponding learnt embedding. To tackle the issue of uncertainty in the reconstruction, we predict multiple reconstructions that are consistent with the input view. This is achieved by learning a probablistic latent space with a novel view-specific \lq{diversity loss}\rq. Thorough quantitative and qualitative analysis is performed to highlight the significance of the proposed approach. We outperform state-of-the-art approaches on the task of single-view 3D reconstruction on both real and synthetic datasets while generating multiple plausible reconstructions, demonstrating the generalizability and utility of our approach. 

\end{abstract}


\section{Introduction}
\label{sec:intro}

\begin{figure*}[h]
\centering
\begin{center}
\includegraphics[width=\linewidth]{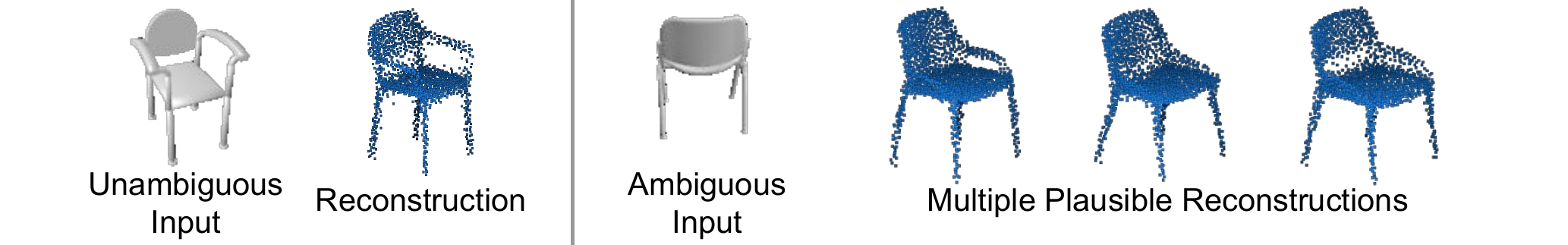}
\end{center}
\caption{Single-view reconstructions for unambiguous and ambiguous input views.}
\label{fig:introduction}
\end{figure*}

Humans can infer the structure of a scene and the shapes of objects within it from limited information. Even for regions that are highly occluded, we are able to guess a number of plausible shapes that could complete the object. Our ability to directly perceive the 3D structure from limited 2D information arises from a strong prior about shapes and geometries that we are familiar with. This ability is central to our perception of the world and the manipulation of objects within it.

Extending the above idea to machines, the ability to infer the 3D structures from single-view images has far-reaching applications in the field of robotics and perception, in tasks such as robot grasping, object manipulation, etc. However, the task is particularly challenging due to the inherent ambiguity  that exists in the reconstructions of occluded images. While the existing data-driven approaches capture the semantic information present in the image to accurately reconstruct corresponding 3D models, it is unreasonable to expect them to predict a single deterministic output for an ambiguous input. An ideal machine would produce multiple solutions when there is uncertainty in the input, while obtaining a deterministic output for images with adequate information (Fig.~\ref{fig:introduction}).  

With the recent advances of deep learning, the problem of 3D reconstruction has largely been tackled with the help of 3D-CNNs that generate a voxelized 3D occupancy grid. However, this representation suffers from sparsity of information, since most of the information needed to perceive the 3D structure is provided by the surface voxels. 3D CNNs are also compute heavy and add considerable overhead during training and inference. To overcome the drawbacks of the voxel representation, recent works have focused on designing neural network architectures and loss formulations to process and predict 3D point clouds ~\cite{qi2017pointnet,qi2017pointnet++,fan2017point}, which consist of points being sampled uniformly on the object's surface. The information-rich encoding and compute-friendly architectures makes it an ideal candidate for 3D shape generation and reconstruction tasks. Hence, we consider point clouds as our 3D representations. 

In this work, we seek to answer two important questions in the task of single-view reconstruction
\begin{enumerate*}[label=\textbf{(\arabic*)}]
    \item Given a two-dimensional image of an object, what is an effective way of inferring an accurate 3D point cloud representation of it?
    \item When the input image is highly occluded, how do we equip the network to generate a set of plausible 3D shapes that are consistent with the input image?
\end{enumerate*}
We achieve the former by first learning a strong prior over all possible 3D shapes with the help of a 3D point cloud auto-encoder. We then train an image encoder to map the input image to this learnt latent space. To address the latter issue, we propose a mechanism to learn a probabilistic distribution in the latent space that is capable of generating multiple plausible outputs from possibly ambiguous input views.

In summary, our contributions in this work are as follows:
\begin{itemize}
    \item We propose a latent-embedding matching setup called 3D-LMNet, to demonstrate the importance of learning a good prior over 3D point clouds for effectively transferring knowledge from the 3D to 2D domain for the task of single-view reconstruction. We thoroughly evaluate various ways of mapping to a learnt 3D latent space.
    \item We present a technique to generate multiple plausible 3D shapes from a single input image to tackle the issue of ambiguous ground truths, and empirically evaluate the effectiveness of this strategy in generating diverse predictions for ambiguous views.
    \item We evaluate 3D-LMNet on real data and demonstrate the generalizability of our approach, which significantly outperforms the state-of-art reconstruction methods for the task of single-view reconstruction. 
\end{itemize}

\section{Related Work}
\label{sec:related_work}

\noindent
\textbf{3D Reconstruction}
\smallskip

With the advent of deep neural network architectures in 2D image generation tasks, the power of convolutional neural nets have been directly transferred to the 3D domain using 3D CNNs. A number of works have revolved around generating voxelized output representations ~\cite{girdhar2016learning,wu2016learning,choy20163d,wu20153d}. Giridhar \etal ~\cite{girdhar2016learning} learnt a joint embedding of 3D voxel shapes and their corresponding 2D images. While the focus of ~\cite{girdhar2016learning} was to learn a vector representation that is generative and predictable at the same time, our aim is to address the problem of transferring the knowledge learnt in 3D to the 2D domain specifically for the task of single-view reconstruction. Additionally, we tackle the rather under-addressed problem of generating multiple plausible outputs that satisfy the given input image. Wu \etal\cite{wu2016learning} used adversarial training in a variational setup for learning more realistic generations. Choy \etal~\cite{choy20163d} trained a recurrent neural network to encode information from more than one input views. Works such as ~\cite{yan2016perspective, tulsiani2017multi} explore ways to reconstruct 3D shapes from 2D projections such as silhouettes and depth maps. Apart from reconstructing shapes from scratch, other reconstruction tasks such as shape completion ~\cite{sharma2016vconv,dai2017shape} and shape deformation ~\cite{yumer2016learning} have also been studied in the voxel domain. But the compute overhead and sparsity of information in voxel formats inspired lines of work that abstracted volumetric information into smaller number of units with the help of the octree data structure~\cite{tatarchenko2017octree, Riegler2017CVPR, hspHane17}.

More recently, Fan \etal~\cite{fan2017point}, introduced frameworks and loss formulations tailored for generating unordered point clouds, and achieved single-view 3D reconstruction results outperforming the volumetric state-of-art approaches~\cite{choy20163d}. While ~\cite{fan2017point} directly predicts the 3D point cloud from 2D images, our approach stresses the importance of first learning a good 3D latent space of point clouds before mapping the 2D images to it. Lin \etal ~\cite{lin2018learning} generated point clouds by fusing depth images and refined them using a projection module. Apart from single-view reconstruction, there is active research in other areas of point cloud analysis including processing ~\cite{deng2018ppfnet, klokov2017escape}, upsampling ~\cite{yu2018pu}, deformation ~\cite{kurenkov2018deformnet}, and generation~\cite{achlioptas2017representation}.

\medskip
\noindent
\textbf{Generating multiple plausible outputs}
\smallskip

While mutliple correct reconstructions can exist for a single input image, most prior works predict deterministic outputs regardless of the information that is available. Rezende \etal~\cite{rezende2016unsupervised} and Fan \etal~\cite{fan2017point} tackle the problem by training a conditional variational auto-encoder~\cite{kingma2013auto,doersch2016tutorial} on 3D shapes conditioned on the input image. In~\cite{fan2017point}, an alternative approach of inducing randomness into the model at the input stage is considered. In 3D-LMNet, we introduce a training regime comprising of sampling a probabilistic latent variable, and optimizing a novel view-specific loss function. Our reconstructions exhibit greater semantic diversity and effectively model the view-specific uncertainty present in the data distribution.

\section{Approach}
\label{sec:approach}

\begin{figure*}[t]
\centering
\begin{center}
\includegraphics[width=\linewidth]{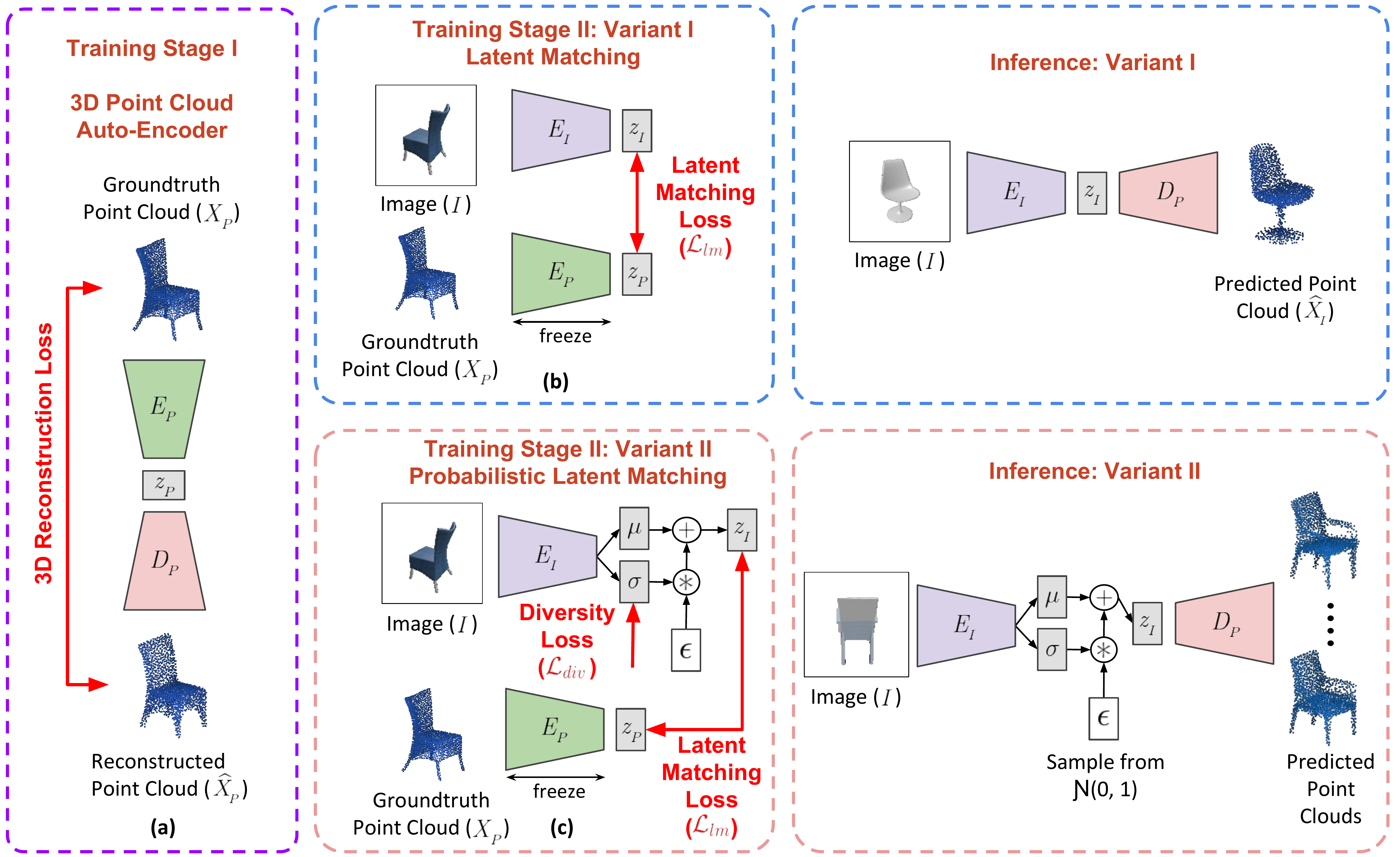}
\end{center}
\caption{Overview of 3D-LMNet. The training pipeline consists of two stages. In Stage I, we learn a latent space $\mathcal{Z}$ for 3D point clouds by training a point cloud auto-encoder ($E_{\!_P}$, $D_{\!_P}$). In Stage II, we train an image encoder $E_{\!_I}$ to map the 2D images to this learnt $\mathcal{Z}$. In a variant of Stage II, we map to $\mathcal{Z}$ in a probabilistic manner so as to infer multiple plausible predictions for a single input image during inference.}
\label{fig:proposed_arch}
\end{figure*}

Our training pipeline consists of two stages as outlined in Fig.~\ref{fig:proposed_arch}. In the first stage, we learn a latent space $\mathcal{Z}\subseteq \mathbb{R}^k$ of 3D point clouds by training a point cloud auto-encoder ($E_{\!_P}$, $D_{\!_P}$). In the second stage, we train an image encoder $E_{\!_I}$ to map the 2D image to this learnt latent space $\mathcal{Z}$. A variant of stage two consists of mapping to $\mathcal{Z}$ in a probabilistic manner so as to infer multiple possible predictions for a single input image during test time. Each of the components is described below in detail.

\subsection{3D Point Cloud Auto-Encoder}

Our goal is to learn a strong prior over the 3D point clouds in the dataset. For this purpose, we train an encoder-decoder network $(E_{\!_P}$, $D_{\!_P})$ that takes in a ground truth point cloud $X_{\!_P}\in \mathbb{R}^{N\times3}$ and outputs a reconstructed point cloud $\widehat{X}_{\!_P}\in \mathbb{R}^{N\times3}$, where $N$ is the number of points in the point cloud (Fig.~\ref{fig:proposed_arch}a). Since a point cloud is an unordered form of representation, we need a network architecture that is invariant to the relative ordering of input points. To enforce this, we choose the architecture of $E_P$ based on PointNet~\cite{qi2017pointnet}, consisting of 1D convolutional layers acting independently on every point in the point cloud $X_{\!_P}$. To achieve order-invariance of point features in the latent space, we apply the maxpool symmetry function to obtain a bottleneck $\mathcal{Z}$ of dimension $k$. The decoder consists of fully-connected layers operating on $\mathcal{Z}$ to produce the reconstructed point cloud $\widehat{X}_{\!_P}$. Since the loss function for optimization also needs to be order-invariant, Chamfer distance between $X_{\!_P}$ and $\widehat{X}_{\!_P}$ is chosen as the reconstruction loss. The loss function is defined as:  

\begin{equation}
    L_{rec} = d_{Chamfer}(X_{\!_P},\widehat{X}_{\!_P}) = \sum_{x\in X_{\!_P}}\min_{\hat{x}\in \widehat{X}_{\!_P}}{||x-\hat{x}||}^2_2 + \sum_{\hat{x}\in \widehat{X}_{\!_P}}\min_{x\in X_{\!_P}}{||x-\hat{x}||}^2_2
\label{eq:chamfer}
\end{equation}

Once the auto-encoder is trained, the next stage consists of training an image encoder to map to this learnt embedding space.

\subsection{Latent Matching}
In this stage, we aim to effectively transfer the knowledge learnt in the 3D point cloud domain to the 2D image domain. We train an image encoder $E_{\!_I}$ that takes in an input image $I$ and outputs a latent image vector $z_{\!_I}$ of dimension $k$ (Variant I, Fig.~\ref{fig:proposed_arch}b). There are two ways of achieving the 3D-to-2D knowledge transfer: 
\begin{enumerate}[label=\textbf{(\arabic*)}]
    \item \textbf{Matching the reconstructions}: We pass $z_{\!_I}$ through the pre-trained point cloud decoder $D_{\!_P}$ to get the predicted point cloud $\widehat{X}_{\!_I}$. The parameters of $D_{\!_P}$ are not updated during this step. Chamfer distance between $\widehat{X}_{\!_I}$ and $X_{\!_P}$ is used as the loss function for optimization. We refer to this variant as "3D-LMNet-Chamfer" in the evaluation section (Sec.~\ref{sec:evaluation}).
    \item \textbf{Matching vectors in the latent $\mathcal{Z}$ space}: The latent representations of the image and corresponding ground truth point cloud are matched. The error is computed between the predicted $z_{\!_I}$ and the ground truth $z_{\!_P}$, obtained via passing $X_{\!_P}$ through the pre-trained point cloud encoder $E_{\!_P}$ (Fig.~\ref{fig:proposed_arch}b). The parameters of $E_{\!_P}$ are not updated during this step. For the latent loss $\mathcal{L}_{lm}$, we experiment with the squared euclidean error ($\mathcal{L}_2(z_{\!_I}-z_{\!_P}) = ||z_{\!_I}-z_{\!_P}||^2_2$) and the least absolute error ($\mathcal{L}_1(z_{\!_I}-z_{\!_P}) = |z_{\!_I}-z_{\!_P}|$) for matching the latent vectors. We refer to these two variants as "3D-LMNet-$\mathcal{L}_2$" and "3DLMNet-$\mathcal{L}_1$" in the evaluation section (Sec.~\ref{sec:evaluation}).
    During inference, we obtain the predicted point cloud by passing the image through $E_{\!_I}$ followed by $D_{\!_P}$. 
\end{enumerate}
In our experiments (detailed in Sec.~\ref{subsec:ablations}), we find that alternative two i.e. matching latent vectors provides substantial improvement over optimizing for the reconstruction loss.

\subsection{Generating Multiple Plausible Outputs}
\label{subsec:variant2}

We propose to handle the uncertainty in predictions by learning a probabilistic distribution in the latent space $\mathcal{Z}$. For every input image $I$ in the dataset, there are multiple settings of the latent variables $z$ for which the model should predict an output that is consistent with $I$. To allow the network to make probabilistic predictions, we formulate the latent representation $z_{\!_1}$ of a specific input image $I_1$ to be a Gaussian random variable, i.e. $z_{\!_1}\thicksim\mathcal{N}(\mu,\sigma^2)$ (Variant II, Fig.~\ref{fig:proposed_arch}c). Similar to Variational Auto-Encoders (VAE)~\cite{kingma2013auto}, we use the "reparameterization trick" to handle stochasticity in the network. The image encoder predicts the mean $\mu$ and standard deviation $\sigma$ of the distribution, and $\epsilon\thicksim\mathcal{N}(0,1)$ is sampled to obtain the latent vector as $z_{\!_1} = \mu+\epsilon\sigma$ (Fig.~\ref{fig:proposed_arch}). However, unlike in the case of conventional VAEs, the mean of the distribution is unconstrained, while the variance is constrained such that meaningful and diverse reconstructions are obtained for a given input image.    

A critical challenge is to obtain a model that can generate diverse but semantically meaningful predictions for an occluded view while retaining the visible semantics. Another challenge is to make highly confident predictions when the input view is informative. To accommodate this, we formulate a fast-decaying loss function that penalizes $\sigma$ for being too far off from zero for unambiguous views, while giving it the liberty to explore the latent space for ambiguous views. 
We term this as the \textit{diversity loss} and define it as follows:
\begin{equation}
    \mathcal{L}_{div} = \big(\sigma-\eta e^{-\frac{(\phi_{i}-\phi_{o})^2}{\delta^2}}\big)^2 
\label{eq:L_div}
\end{equation}

where, $\phi_{i}$ is the azimuth angle of the input image $I$, $\phi_{o}$ is the azimuth angle of maximum occlusion view, and $\delta$ determines the rate of decay. $\eta$ controls the magnitude of standard deviation $\sigma$. The above formulation can easily be extended to cases where multiple highly occluded views are present by considering a mixture of Gaussians. 

The joint optimization loss function is a combination of the latent matching loss $\mathcal{L}_{lm}$ and the diversity loss $\mathcal{L}_{div}$:
\begin{equation}
    \mathcal{L} = \mathcal{L}_{lm} + \lambda \mathcal{L}_{div} 
\end{equation}
where $\lambda$ is the weighing factor.
During inference, the model is capable of generating diverse predictions when $\epsilon$ is varied. Note that pose information is not used during inference. 

\begin{table}[t]
\centering
\begin{center}
\begin{tabular}{|c|c|cccc|}
\hline
Metric  & AE & Baseline & \begin{tabular}[c]{@{}c@{}}3D-LMNet\\ Chamfer\end{tabular} & \begin{tabular}[c]{@{}c@{}}3D-LMNet\\ $\mathcal{L}_2$\end{tabular} & \begin{tabular}[c]{@{}c@{}}3D-LMNet\\ $\mathcal{L}_1$\end{tabular} \\ \hline\hline
Chamfer  &  4.46  &  5.78    & 5.99  & 5.54  & \textbf{5.40}  \\ \hline  
EMD &  6.53  &  9.20    & 7.82  & 7.20   & \textbf{7.00}  \\ \hline
\end{tabular}
\end{center}
\caption{A comparison of the baseline and different variants of 3D-LMNet for the task of 3D reconstruction on ShapeNet~\cite{chang2015shapenet}. All metrics are scaled by 100.}
\label{tab:rec_ablations}
\end{table}

\begin{table}[t]
\centering
\begin{center}
\begin{tabular}{|c|ccc|}
\hline
Metric & \begin{tabular}[c]{@{}c@{}}3D-LMNet\\ Chamfer\end{tabular} & \begin{tabular}[c]{@{}c@{}}3D-LMNet\\ $\mathcal{L}_2$\end{tabular} & \begin{tabular}[c]{@{}c@{}}3D-LMNet\\ $\mathcal{L}_1$\end{tabular} \\ \hline\hline
$\mathcal{L}_2$ &  56.7 & \textbf{1.32}  & 1.38  \\ \hline 
$\mathcal{L}_1$ &  14.02 & 1.34  & \textbf{1.29}  \\ \hline
\end{tabular}
\end{center}
\caption{A comparison of latent matching errors for different variants of 3D-LMNet on ShapeNet~\cite{chang2015shapenet}. All metrics are scaled by 0.01.}
\label{tab:lm_ablations}
\end{table}

\subsection{Implementation Details}
In the point cloud auto-encoder, the encoder consists of five 1D convolutional layers with $[64,128,128,256]$ filters, ending with a bottleneck layer of dimension 512. We choose maxpool function as the symmetry operation. The decoder consists of three fully-connected layers of size $[256,256,N\times3]$, where $N$ is the number of points predicted by our network. We set $N$ to be 2048 in all our experiments. We use the ReLU non-linearity and batch-normalization at all layers of the auto-encoder. The image encoder is a 2D convolutional neural network that maps the input image to the 512-dimensional latent vector. We use the Adam optimizer with a learning rate of $0.00005$ and a minibatch size of 32. Network architectures for all components in our proposed framework are provided in the supplementary material. Codes are available at \url{https://github.com/val-iisc/3d-lmnet}.

\section{Experiments}
\label{sec:evaluation}

\noindent\textbf{Dataset}: We train all our networks on synthetic models from the ShapeNet~\cite{chang2015shapenet} dataset. We use the same $80\%-20\%$ train/test split provided by \cite{choy20163d} consisting of models from 13 different categories, so as to be comparable with the previous works. 

\noindent\textbf{Evaluation Methodology}: We report both the Chamfer Distance (Eqn.~\ref{eq:chamfer}) as well as the Earth Mover's Distance (or EMD) computed on 1024 randomly sampled points in all our evaluations. EMD between two point sets $X_{\!_P}$ and $\widehat{X}_{\!_P}$ is given by:
\begin{equation}
\label{eq:emd}
    d_{EMD}(X_{\!_P},\widehat{X}_{\!_P})=\min_{\phi:X_{\!_P}\rightarrow \widehat{X}_{\!_P}}\sum_{x\in X_{\!_P}}||x-\phi(x)||_2
\end{equation}
where $\phi:X_{\!_P}\rightarrow \widehat{X}_{\!_P}$ is a bijection. We use an approximate. For computing the metrics, we renormalize both the ground truth and predicted point clouds within a bounding box of length 1 unit. Since PSGN~\cite{fan2017point} outputs are non-canonical, we align their predictions to the canonical ground truth frame by using pose metadata available in the evaluation datasets. Additionally, we apply the iterative closest point algorithm (ICP)~\cite{besl1992method} on the ground truth and predicted point clouds for finer alignment. 

\begin{figure*}[t]
\centering
\begin{center}
    \includegraphics[width=\linewidth]{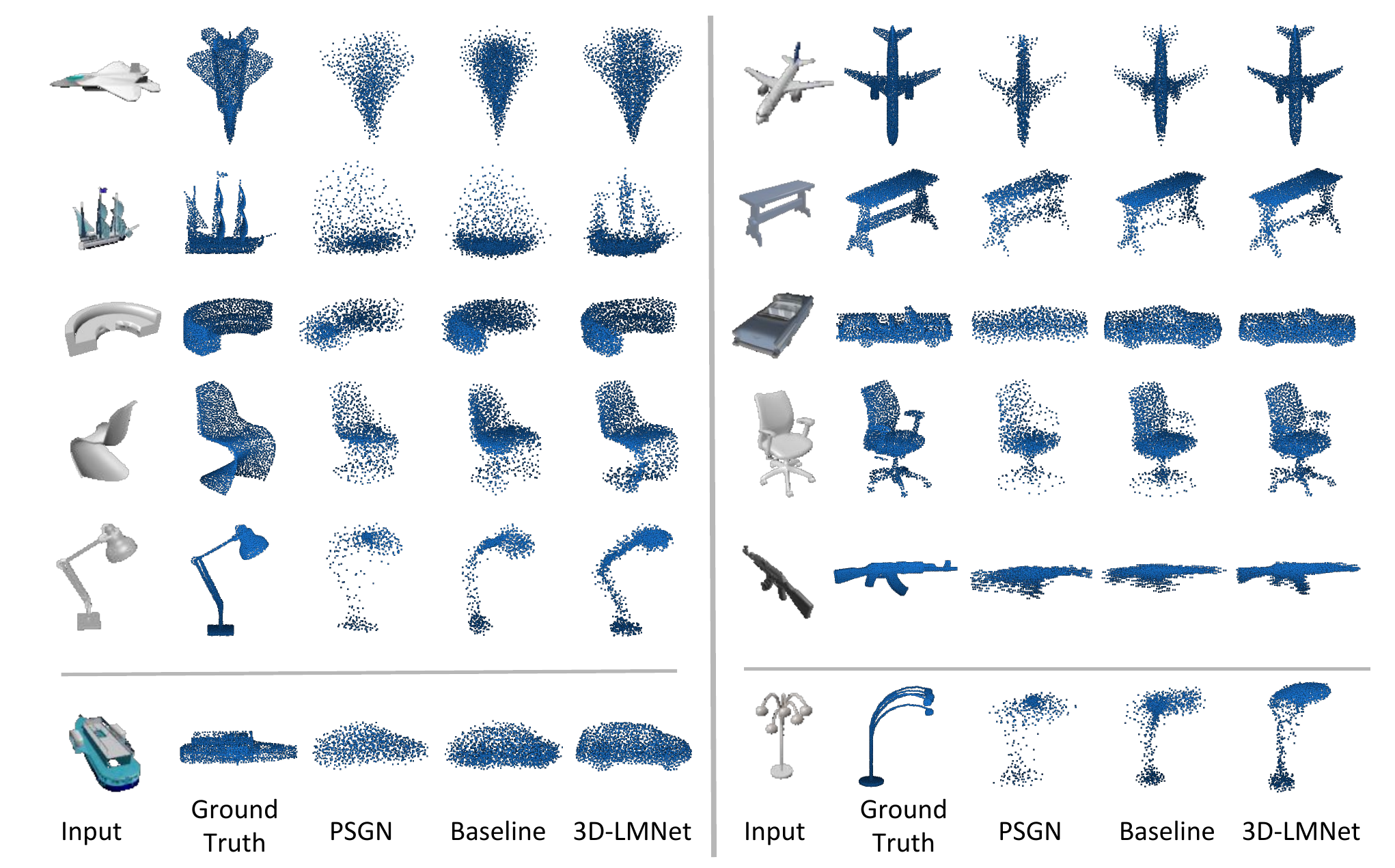}
\end{center}
\caption{Qualitative results on ShapeNet~\cite{chang2015shapenet}. Compared to PSGN~\cite{fan2017point} and the baseline, we are better able to capture the overall shape and finer details present in the input image. While clusters and outlier points are present in PSGN and baseline reconstructions, we obtain more uniformly distributed points. The bottom row presents failure cases for our approach. Note that PSGN predicts 1024 points, while the baseline and 3D-LMNet predict 2048 points.}
\label{fig:shapenet_comparison}
\end{figure*}

\subsection{Empirical Evaluation on ShapeNet}
\label{subsec:ablations}
We study the framework presented and evaluate each of the components in the training pipeline. To show the advantage of our latent matching procedure over direct 2D-3D training, we train a baseline which consists of an encoder-decoder network that is trained end-to-end, using reconstruction loss on the generated point cloud. We employ the same network architecture as the one used for latent matching experiments. To measure the performance of the baseline and all variants of our model, we use the validation split provided by~\cite{choy20163d} for reporting the Chamfer and EMD metrics. Table~\ref{tab:rec_ablations} shows the comparison between the baseline and three variants of loss formulation in our latent matching setup. We also report the auto-encoder reconstruction scores, which serve as an upper bound on the performance of latent matching. We observe that the latent matching variants of 3D-LMNet outperform the baseline in both Chamfer and EMD. Amongst the variants, we observe that trivially optimizing for Chamfer loss leads to worse results, whereas training with losses directly operating on the latent space results in lower reconstruction errors. We also see that the $\mathcal{L}_1$ loss formulation performs better both in terms of Chamfer and EMD metrics. Additionally, we also report the latent matching errors for different variants of 3D-LMNet in Table~\ref{tab:lm_ablations}. We observe that more accurate latent matching (characterized by lower $\mathcal{L}_1$ and $\mathcal{L}_2$ errors), results in lower reconstruction errors as well (Table~\ref{tab:rec_ablations}). Category-wise metrics for all the variants are provided in the supplementary material.

\subsection{Comparison with other methods on ShapeNet and Pix3D}
 We compare our 3D-LMNet-$\mathcal{L}_1$ model with PSGN~\cite{fan2017point} on the synthetic ShapeNet dataset~\cite{chang2015shapenet} and the more recent Pix3D dataset~\cite{pix3d} to test for generalizability on real world data. Since~\cite{fan2017point} establishes that point cloud based approach significantly outperforms the state-of-art voxel based approaches, we do not show any comparison against them.

\smallskip 
\noindent\textbf{ShapeNet} Table~\ref{tab:sota_shapenet} shows the comparison between PSGN~\cite{fan2017point}, the baseline and our $\mathcal{L}_1$ latent matching variant on the validation split provided by ~\cite{choy20163d}. We outperform PSGN in \textcolor{red}{8} out of 13 categories in Chamfer and \textcolor{red}{10} out of 13 categories in the EMD metric, while also having lower overall mean scores. It is worth noting that we achieve state-of-the-art performance in both metrics despite the fact that our network has half the number of trainable parameters in comparison to PSGN, while predicting point clouds with double the resolution. A lower EMD score also correlates with better visual quality and encourages points to lie closer to the surface~\cite{achlioptas2017representation,yu2018pu}. Qualitative comparison is shown in Fig. ~\ref{fig:shapenet_comparison}. Compared to PSGN~\cite{fan2017point} and the baseline, we are better able to capture the overall shape and finer details present in the input image. Note that both the other methods have clustered points and outlier points while our reconstructions are more uniformly distributed. We also present two failure cases of our approach in Fig.~\ref{fig:shapenet_comparison} (bottom row). Interestingly, we observe that in some cases, latent matching incorrectly maps an image to a similar looking object of different category, leading to good-looking but incorrect reconstructions. Fig.~\ref{fig:shapenet_comparison} shows a vessel being mapped to a car of similar shape. Another common failure case is the absence of finer details in the reconstructions. However, other approaches also have this drawback. 

\begin{table}[]
\centering
\begin{center}
\begin{tabular}{|c|ccc|ccc|}
\hline
\multirow{2}{*}{Category}      & \multicolumn{3}{c|}{Chamfer} & \multicolumn{3}{c|}{EMD} \\ \cline{2-7} 
& Baseline   & PSGN~\cite{fan2017point}   & 3D-LMNet  & Baseline  & PSGN~\cite{fan2017point}  & 3D-LMNet \\ \hline\hline
airplane   &    3.61    &   3.74     &  \textbf{3.34}    & 7.42    & 6.38    &  \textbf{4.77}    \\
bench      &    4.70    &   4.63     &  \textbf{4.55}    & 5.66    & 5.88    &  \textbf{4.99}    \\
cabinet    &    7.42    &   6.98     &  \textbf{6.09}    & 9.58    & \textbf{6.04}    &  6.35    \\
car        &    4.67    &   5.20     &  \textbf{4.55}    & 4.74    & 4.87    &  \textbf{4.10}    \\
chair      &    6.51    &   \textbf{6.39}     &  6.41    & 8.99    & 9.63    &  \textbf{8.02}    \\
lamp       &    7.32    &   \textbf{6.33}     &  7.10    & 20.96   & 16.17   &  \textbf{15.80}   \\
monitor    &    6.76    &   \textbf{6.15}     &  6.40    & 9.18    & 7.59    &  \textbf{7.13}    \\
rifle      &    2.99    &   2.91     &  \textbf{2.75}    & 9.30    & 8.48    &  \textbf{6.08}    \\
sofa       &    6.11    &   6.98     &  \textbf{5.85}    & 6.40    & 7.42    &  \textbf{5.65}    \\
speaker    &    9.05    &   8.75     &  \textbf{8.10}    & 11.29   & \textbf{8.70}    &  9.15    \\
table      &    6.16    &   \textbf{6.00}     &  6.05    & 9.51    & 8.40    &  \textbf{7.82}    \\
telephone  &    5.13    &   \textbf{4.56}     &  4.63    & 8.64    & \textbf{5.07}    &  5.43    \\
vessel     &    4.70    &   4.38     &  \textbf{4.37}    & 7.88    & 6.18    &  \textbf{5.68}    \\ \hline
\textbf{mean} & 5.78    &    5.62    &  \textbf{5.40}    & 9.20    & 7.75    &  \textbf{7.00}    \\ \hline
\end{tabular}
\end{center}
\caption{Single view reconstruction results on ShapeNet~\cite{chang2015shapenet}. The metrics are computed on 1024 points after performing ICP alignment with the ground truth point cloud. All metrics are scaled by 100.}
\label{tab:sota_shapenet}
\end{table}

\begin{figure*}
\centering
\begin{center}
    \includegraphics[width=\linewidth]{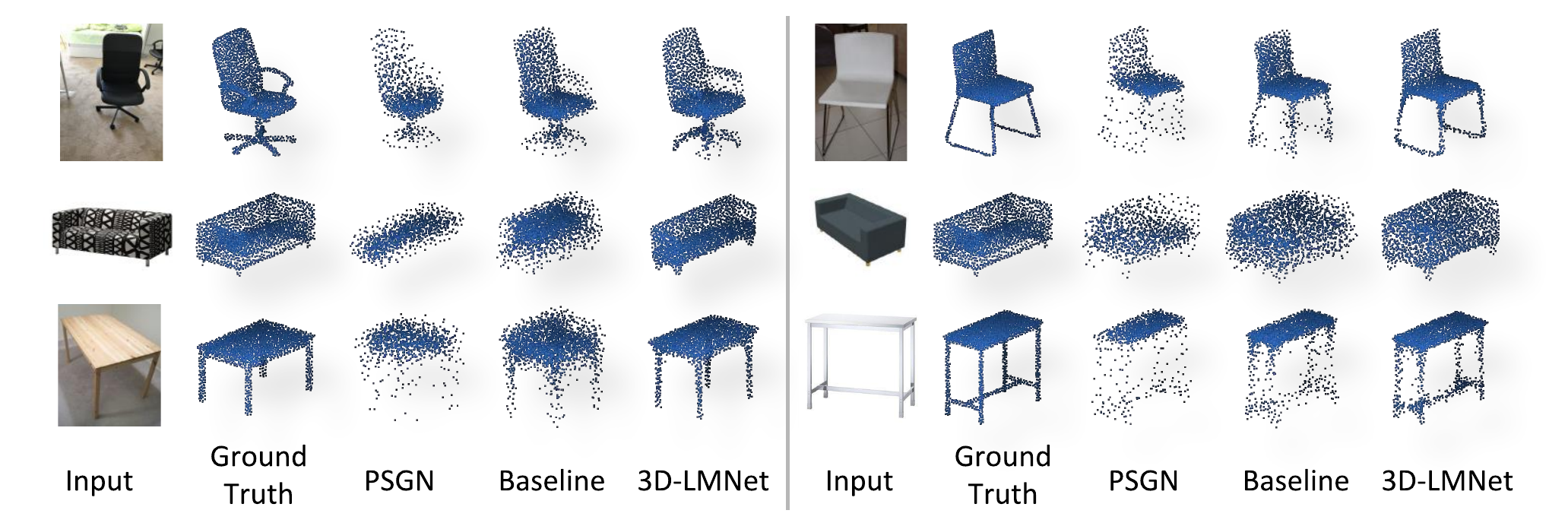}
\end{center}
\caption{Qualitative results on the real-world Pix3D dataset~\cite{pix3d}. The learnt 3D prior enables our method to generate meaningful reconstructions on real data, while both PSGN~\cite{fan2017point} and the baseline fail to generalize well to the real data distribution. PSGN predicts 1024 points, while the baseline and 3D-LMNet predict 2048 points.}
\label{fig:pix3d_comparison}
\end{figure*}

\smallskip
\noindent\textbf{Pix3D} For testing the generalizability of our approach on real-world datasets, we evaluate the performance of our method on the Pix3D dataset~\cite{pix3d}. It consists of a large collection of real images and their corresponding metadata such as masks, ground truth CAD models and pose. We evaluate our trained model on categories that co-occur in the synthetic training set and exclude images having occlusion and truncation from the test set, as is done in the original paper~\cite{pix3d}. We crop the images to center-position the object of interest and mask the background using the provided information. We report the results of this evaluation in Table \ref{tab:sota_pix3d}. Evidently, we outperform PSGN and the baseline by a large margin in both Chamfer as well as EMD metrics, demonstrating the effectiveness of our approach on real data. Fig.~\ref{fig:pix3d_comparison} shows sample reconstructions on this dataset. Our proposed method is able to generalize well to the real dataset while both PSGN and the baseline struggle to generate meaningful reconstructions.

\begin{table}[]
\centering
\begin{center}
\begin{tabular}{|c|ccc|ccc|}
\hline
\multirow{2}{*}{Category}      & \multicolumn{3}{c|}{Chamfer} & \multicolumn{3}{c|}{EMD} \\ \cline{2-7} 
& Baseline   & PSGN~\cite{fan2017point}   & 3D-LMNet  & Baseline  & PSGN~\cite{fan2017point}  & 3D-LMNet \\ \hline\hline
chair     & 7.52  & 8.05  & \textbf{7.35} & 11.17 & 12.55 & \textbf{9.14}  \\
sofa      & 8.65  & 8.45  & \textbf{8.18} & 8.87  & 9.16  & \textbf{7.22}  \\
table     & 11.23 & \textbf{10.82} & 11.20 & 15.71 & 15.16 & \textbf{12.73} \\ \hline
\textbf{mean} & 9.13  & 9.11  & \textbf{8.91} & 11.92 & 12.29 & \textbf{9.70} \\ \hline
\end{tabular}
\end{center}
\caption{Single view reconstruction results on the real world Pix3D dataset~\cite{pix3d}. The metrics are computed on 1024 points after performing ICP alignment with the ground truth point cloud. All metrics are scaled by 100.}
\label{tab:sota_pix3d}
\end{table}

\subsection{Generating multiple plausible outputs}

\begin{figure*}
\centering
\begin{center}
    \includegraphics[width=\linewidth]{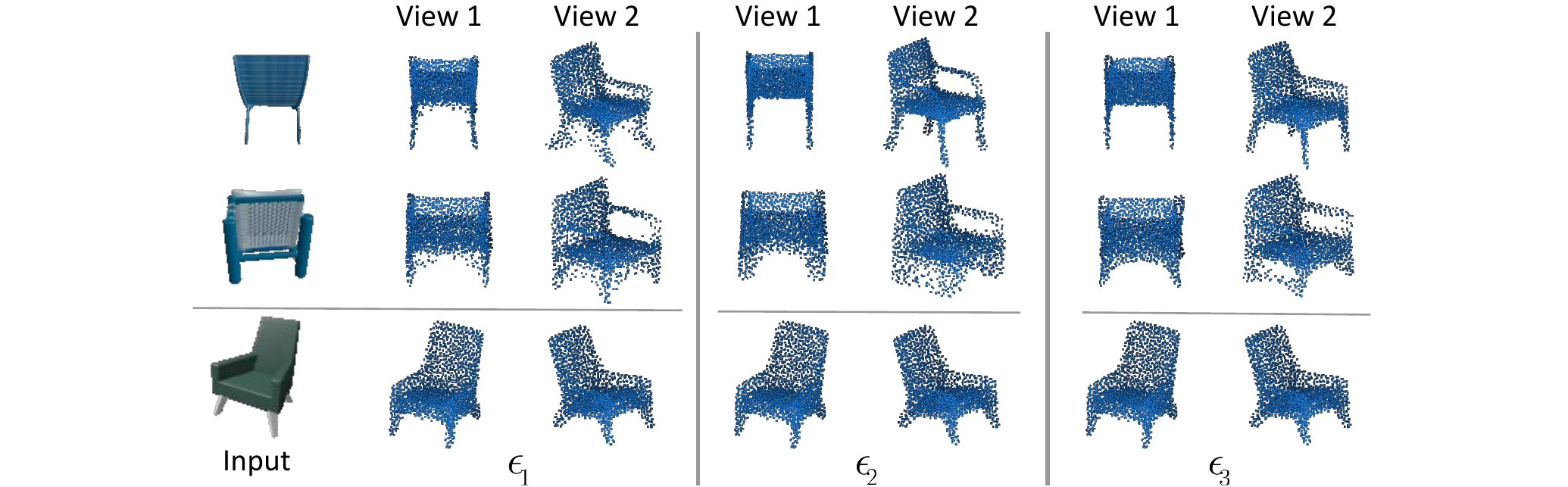}
\end{center}
\caption{Qualitative results for probabilistic latent matching. Rows 1 and 2 depict multiple reconstructions for ambiguous input views, obtained by sampling $\epsilon$ from $\mathcal{N}(0,1)$. Row 3 depicts the minimal influence $\epsilon$ has on reconstructions from informative views. Reconstruction results are shown from two different viewing angles for each $\epsilon$ so as to highlight the correspondence with the input image.}
\label{fig:vae_results}
\end{figure*}

We evaluate the probabilistic training regime (Variant II, Fig.~\ref{fig:proposed_arch}c) described in Sec.~\ref{subsec:variant2} for the task of generating multiple plausible outputs for a single input image. We train $E_{\!_I}$ on objects from the chair category, and set $\phi_o$ and $\delta$ in Eqn.~\ref{eq:L_div} to $180^{\circ}$ and $20^{\circ}$ respectively. For chairs, $\phi_o$ of $180^{\circ}$ corresponds to a perfect back-view having maximum occlusion. For comparison, we also train a model without the diversity loss (Variant I, Fig.~\ref{fig:proposed_arch}b). Quantitatively, Variant II compares favourably to Variant I in terms of both Chamfer (Variant II - $6.45$ vs Variant I - $6.48$) and EMD errors (Variant II - $8.04$ vs Variant I - $8.1$), while also effectively handling uncertainty. Qualitative results for Variant II are shown in Fig.~\ref{fig:vae_results}. We observe that for different values of the sampling variable $\epsilon\thicksim\mathcal{N}(0,1)$, we obtain semantically different reconstructions which are consistent with the input image for ambiguous views. We observe variations like presence and absence of handles, different leg structures, hollow backs, etc in the reconstructions. On the other hand, the value of $\epsilon$ has minimal influence over the reconstructions for unambiguous views.

\section{Conclusion}
\label{sec:conclusion}

In this paper, we highlighted the importance of learning a rich latent representation of 3D point clouds for the task of single-view 3D reconstruction. We empirically evaluated various loss formulations to effectively map to the learned latent space. We also presented a technique to tackle the inherent ambiguity in 3D shape prediction from a single image by introducing a probabilistic training scheme in the image encoder, thereby obtaining multiple plausible 3D generations from a single input image. Quantitative and qualitative evaluation on the single-image 3D reconstruction task on synthetic and real datasets show that the generated point clouds are more accurate and realistic in comparison to the current state-of-art reconstruction methods.

\bibliography{biblio}

\section*{\centering \LARGE{Supplementary Material}}

\setcounter{section}{0}
\setcounter{table}{0}
\setcounter{figure}{0}
\setcounter{equation}{0}

\section{Training Dataset Details}
We train all our networks on synthetic models from the ShapeNet~\cite{chang2015shapenet} dataset. We use the same $80\%$-$20\%$ train/test split provided by~\cite{choy20163d} consisting of models from 13 different categories, so as to be comparable with the previous works. We use the input images provided by~\cite{choy20163d}, where each model is pre-rendered from 24 different azimuth angles. We crop the images to $128\times 128$ resolution before passing it through our network. For generating the ground truth point cloud, we uniformly sample 16,384 points on the mesh surface using farthest point sampling.

\section{Network Architectures}
We provide network architecture details for the point cloud and image encoders and the common decoder in Tables~\ref{tab:point_cloud_encoder_architecture},\ref{tab:image_encoder_architecture} and \ref{tab:point_cloud_decoder_architecture}. It should be noted that 3D-LMNet has a total of 22.7M parameters, while PSGN~\cite{fan2017point} has nearly double the number with 42.9M parameters.

\begin{table}[h]
\centering
\begin{tabular}{|c|c|c|c|c|}
\hline
S.No. & Layer    & Filter Size & Output Size & Params \\ \hline\hline
1     & conv    & 1x1         & 2048x64     & 0.4K   \\
2     & conv    & 1x1         & 2048x128    & 8.6K   \\
3     & conv    & 1x1         & 2048x128    & 16.8K  \\
4     & conv    & 1x1         & 2048x256    & 33.5K  \\
5     & conv    & 1x1         & 2048x512    & 132.6K \\
6     & maxpool & -           & 512         & 0      \\ \hline
\end{tabular}
\smallskip
\caption{Point Cloud Encoder Architecture}
\label{tab:point_cloud_encoder_architecture}
\end{table}

\begin{table}[h]
\centering
\begin{tabular}{|c|c|c|c|}
\hline
S.No. & Layer   & \begin{tabular}[c]{@{}c@{}}Filter Size/\\ Stride\end{tabular} & Output Size  \\ \hline\hline
1     & conv   & 3x3/1                                                         & 64x64x32   \\
2     & conv   & 3x3/1                                                         & 64x64x32   \\
3     & conv   & 3x3/2                                                         & 32x32x64     \\
4     & conv   & 3x3/1                                                         & 32x32x64     \\
5     & conv   & 3x3/1                                                         & 32x32x64     \\
6     & conv   & 3x3/2                                                         & 16x16x128    \\
7     & conv   & 3x3/1                                                         & 16x16x128    \\
8     & conv   & 3x3/1                                                         & 16x16x128    \\
9     & conv   & 3x3/2                                                         & 8x8x256    \\
10    & conv   & 3x3/1                                                         & 8x8x256    \\
11    & conv   & 3x3/1                                                         & 8x8x256    \\
16    & conv   & 5x5/2                                                         & 4x4x512      \\
17    & linear & -                                                             & 128          \\ \hline
\end{tabular}
\smallskip
\caption{Image Encoder Architecture}
\label{tab:image_encoder_architecture}
\end{table}

\begin{table}[t]
\centering
\begin{tabular}{|c|c|c|}
\hline
S.No. & Layer   & Output Size\\ \hline\hline
1     & linear & 256         \\
2     & linear & 256         \\
3     & linear & 1024*3      \\ \hline
\end{tabular}
\smallskip
\caption{Decoder Architecture}
\label{tab:point_cloud_decoder_architecture}
\end{table}

\section{Quantitative Comparison of 3D-LMNet Variants on ShapeNet}
We report the category-wise Chamfer and EMD error metrics for all our latent matching variants on the validation split provided by~\cite{choy20163d} for the ShapeNet dataset~\cite{chang2015shapenet} in Table~\ref{tab:sota_shapenet}. Our latent matching approaches (3D-LMNet-$\mathcal{L}_1$ and $\mathcal{L}_2$) significantly outperform the network trained directly with Chamfer loss (3D-LMNet-Chamfer). 3D-LMNet-$\mathcal{L}_1$ is better in all categories in Chamfer scores, and all but one category in terms of EMD scores.

\begin{table}[t]
\centering
\tabcolsep=4pt\relax
\small
\begin{center}
\begin{tabular}{|c|ccc|ccc|}
\hline
\multirow{2}{*}{Category}      & \multicolumn{3}{c|}{Chamfer} & \multicolumn{3}{c|}{EMD} \\ \cline{2-7} 
& \begin{tabular}[c]{@{}c@{}}3D-LMNet\\ Chamfer\end{tabular} & \begin{tabular}[c]{@{}c@{}}3D-LMNet\\ $\mathcal{L}_2$\end{tabular}   & \begin{tabular}[c]{@{}c@{}}3D-LMNet\\ $\mathcal{L}_1$\end{tabular}   & \begin{tabular}[c]{@{}c@{}}3D-LMNet\\ Chamfer\end{tabular}  & \begin{tabular}[c]{@{}c@{}}3D-LMNet\\ $\mathcal{L}_2$\end{tabular}  & \begin{tabular}[c]{@{}c@{}}3D-LMNet\\ $\mathcal{L}_1$\end{tabular} \\ \hline\hline
airplane      & 4.47 & 3.39 & \textbf{3.34} & 7.35  & 4.81          & \textbf{4.77}  \\
bench         & 5.03 & 4.74 & \textbf{4.55} & 5.38  & 5.17          & \textbf{4.99}  \\
cabinet       & 6.76 & 6.26 & \textbf{6.09} & 7.03  & 6.73          & \textbf{6.35}  \\
car           & 4.70 & 4.61 & \textbf{4.55} & 4.31  & 4.20          & \textbf{4.10}  \\
chair         & 6.72 & 6.54 & \textbf{6.41} & 8.16  & 8.11          & \textbf{8.02}  \\
lamp          & 8.31 & 7.28 & \textbf{7.10} & 17.21 & 16.03         & \textbf{15.80} \\
monitor       & 6.96 & 6.65 & \textbf{6.40} & 7.66  & 7.53          & \textbf{7.13}  \\
rifle         & 3.03 & 2.79 & \textbf{2.75} & 6.67  & \textbf{6.06} & 6.08           \\
sofa          & 6.20 & 6.00 & \textbf{5.85} & 5.97  & 5.80          & \textbf{5.65}  \\
speaker       & 8.77 & 8.33 & \textbf{8.10} & 9.20  & 9.61          & \textbf{9.15}  \\
table         & 6.59 & 6.16 & \textbf{6.05} & 8.34  & 7.95          & \textbf{7.82}  \\
telephone     & 5.62 & 4.87 & \textbf{4.63} & 7.50  & 5.79          & \textbf{5.43}  \\
vessel        & 4.76 & 4.45 & \textbf{4.37} & 6.92  & 5.84          & \textbf{5.68}  \\ \hline
\textbf{mean} & 5.99 & 5.54 & \textbf{5.40} & 7.82  & 7.20          & \textbf{7.00} \\ \hline
\end{tabular}
\end{center}
\caption{Category-wise 3D reconstruction metrics for different latent matching variants of 3D-LMNet on the ShapeNet dataset~\cite{chang2015shapenet}. All metrics are scaled by 100.}
\label{tab:sota_shapenet}
\end{table}

\section{Reconstructions on ShapeNet}

\begin{figure*}[t]
\centering
\begin{center}
    \includegraphics[width=\linewidth]{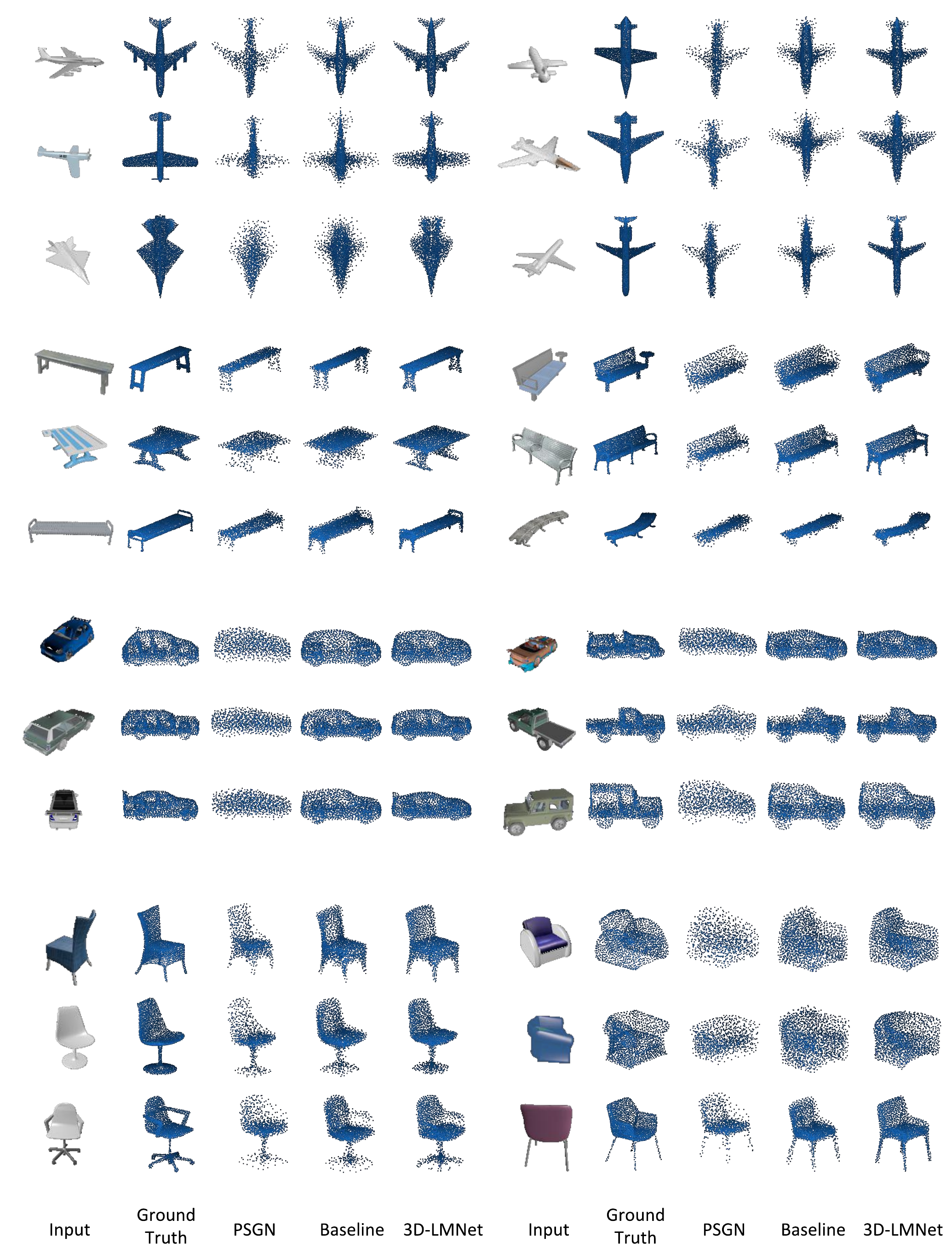}
\end{center}
\caption{Reconstructions on ShapeNet. 3D reconstructions on randomly sampled input images from the validation set of ShapeNet. Note that although the baseline reconstructions for cars obtain a good shape, the points are unevenly distributed which results in high EMD error metrics (main text Table~\ref{tab:sota_shapenet}). On the other hand, 3D-LMNet reconstructions are well distributed and obtain lower EMD error metrics. Results best viewed zoomed.}
\label{fig:shapenet_reconstructions}
\end{figure*}

\begin{figure*}[t]
\centering
\begin{center}
    \includegraphics[width=\linewidth]{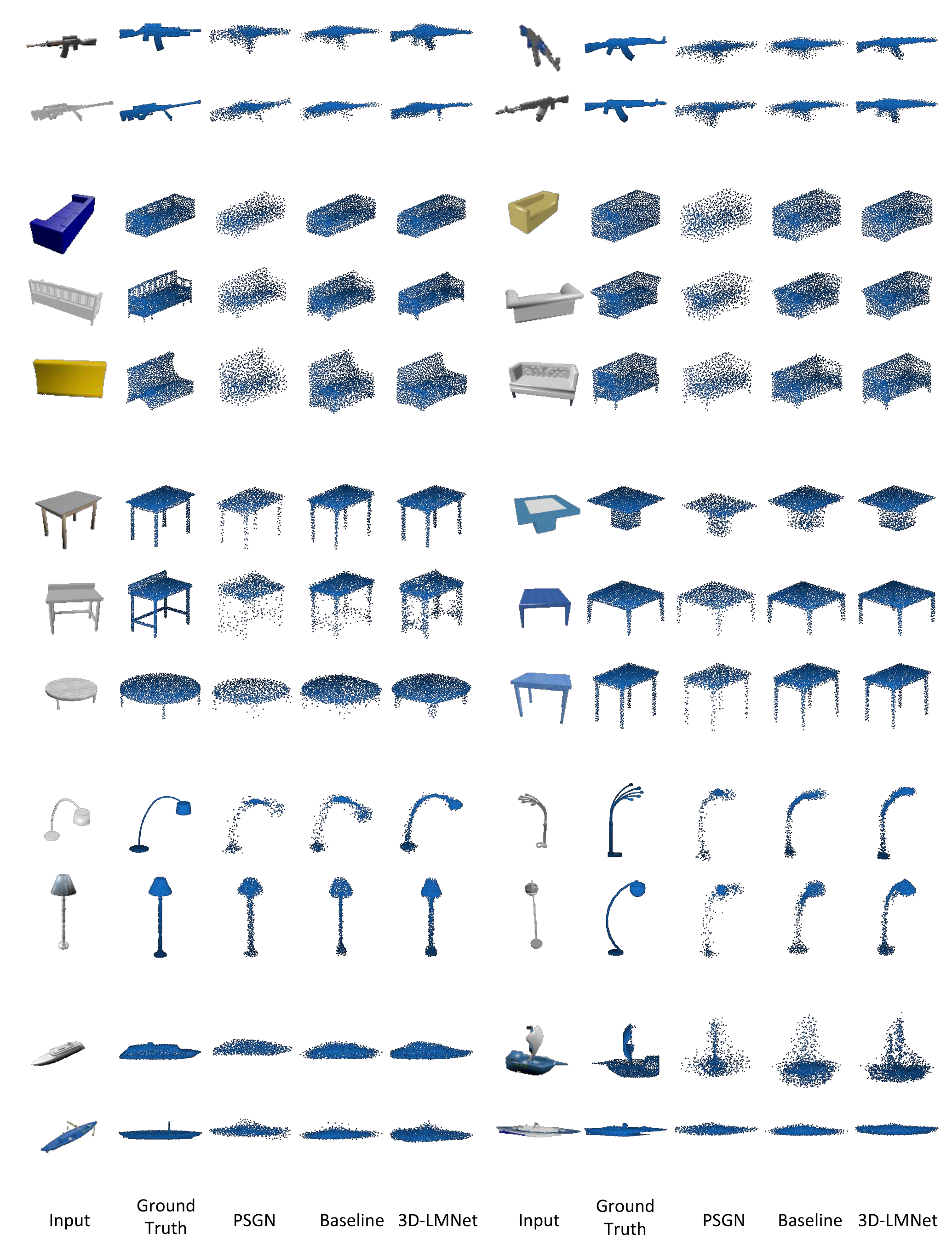}
\end{center}
\caption{Reconstructions on ShapeNet. 3D reconstructions on randomly sampled input images from the validation set of ShapeNet. Results best viewed zoomed.}
\label{fig:shapenet_reconstructions_2}
\end{figure*}

Qualitative comparison with state-of-art and baseline for single-view reconstruction on ShapeNet validation set are provided in Figs.~\ref{fig:shapenet_reconstructions} and~\ref{fig:shapenet_reconstructions_2}. Note that the samples are randomly selected. 

\section{Reconstructions on Pix3D}

\begin{figure*}[t]
\centering
\begin{center}
    \includegraphics[width=\linewidth]{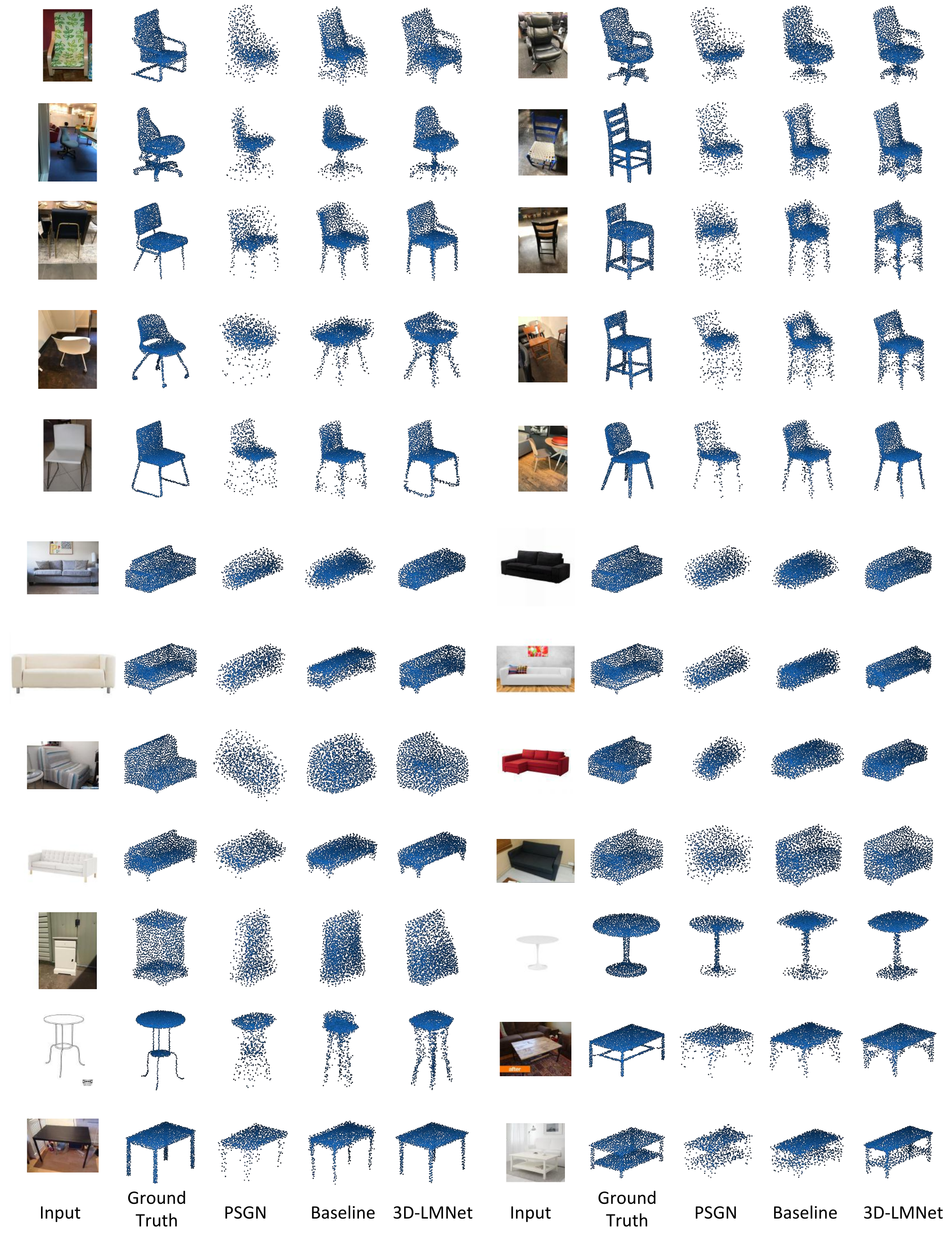}
\end{center}
\caption{Reconstructions on Pix3d. 3D reconstructions on randomly sampled input images from Pix3D. Results best viewed zoomed.}
\label{fig:pix3d_reconstructions}
\end{figure*}

Qualitative comparison with state-of-art and baseline for single-view reconstruction on the real-world Pix3D dataset are shown in Fig.~\ref{fig:pix3d_reconstructions}. Note that the samples are randomly selected. 

\section{Generating Multiple Plausible Outputs}

\begin{figure*}[t]
\centering
\begin{center}
    \includegraphics[width=\linewidth]{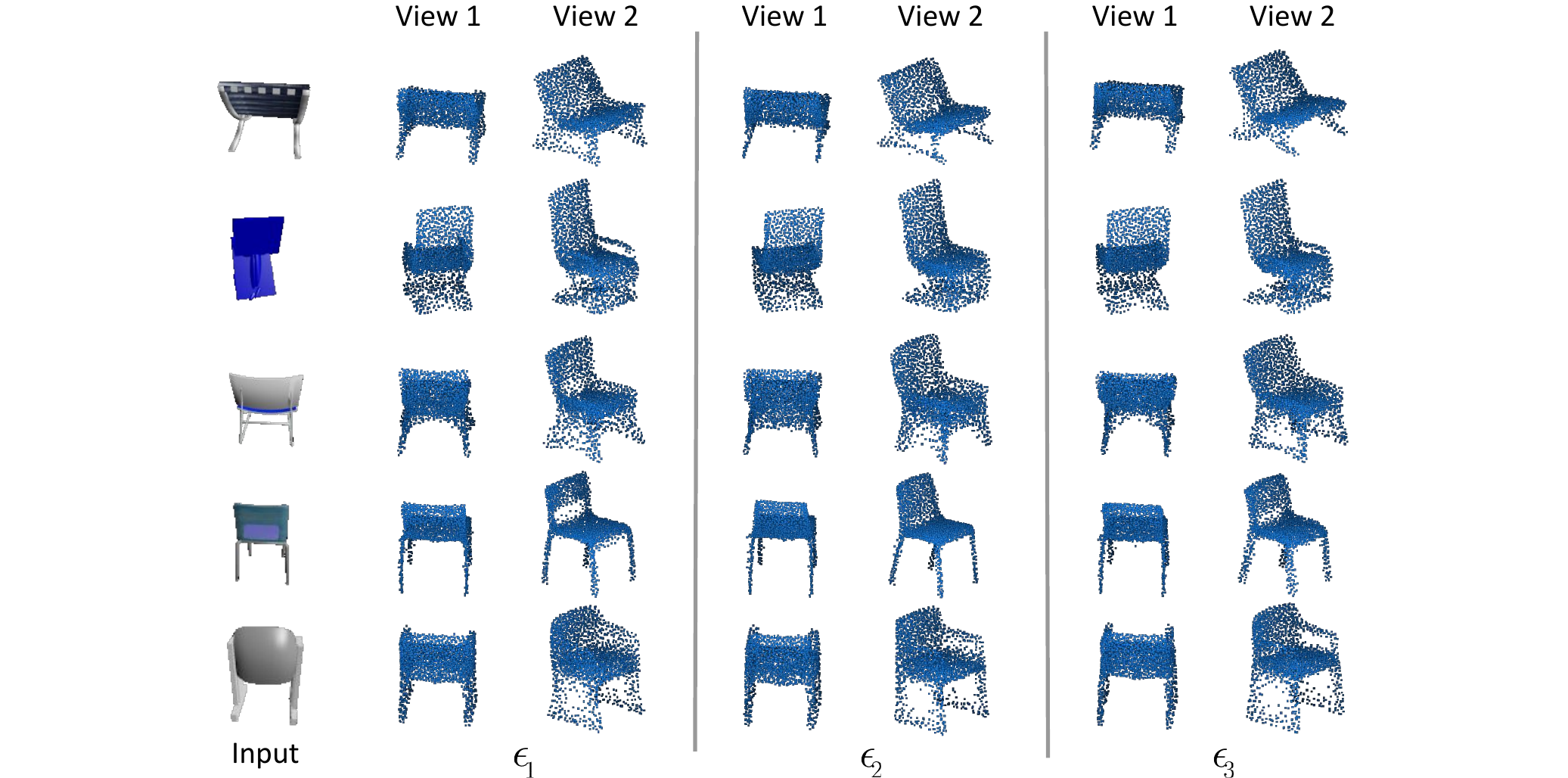}
\end{center}
\caption{Qualitative results for probabilistic latent matching. Multiple reconstructions for ambiguous input views are obtained by sampling $\epsilon$. Reconstruction results are shown from two different viewing angles for each $\epsilon$ so as to highlight the correspondence with the input image.}
\label{fig:vae}
\end{figure*}

We provide more examples for the probabilistic latent matching scheme explained in the paper in Fig.~\ref{fig:vae}. We notice variations in legs, handles and back of the chair models.

\section{Auto-Encoder Results}

\subsection{Reconstructions}
\begin{figure*}[t]
\centering
\begin{center}
    \includegraphics[width=\linewidth]{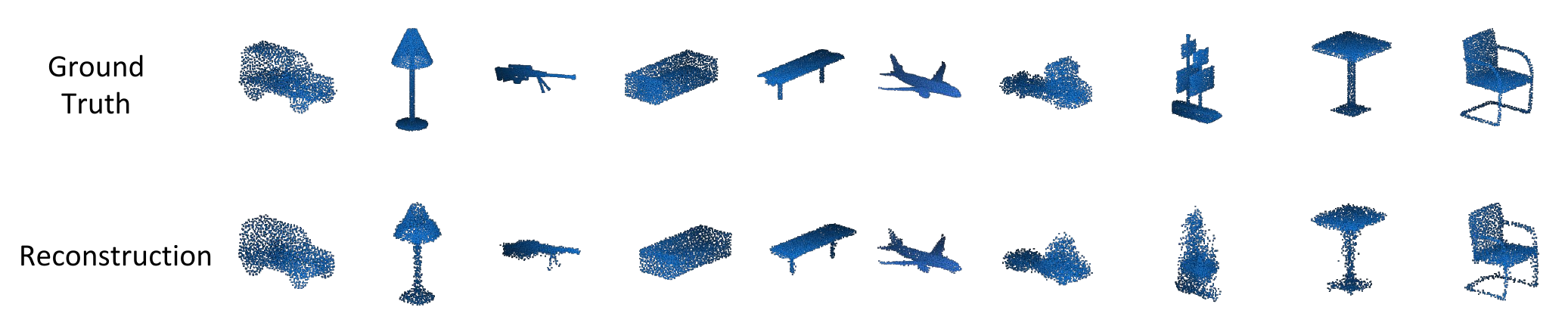}
\end{center}
\caption{Auto-Encoder Reconstructions}
\label{fig:ae_reconstructions}
\end{figure*}

3D point cloud reconstruction results are shown in Fig.~\ref{fig:ae_reconstructions}. The reconstructions are very similar to the ground truth point clouds in appearance and spread.

\subsection{Latent Space Interpolations}
\begin{figure*}[t]
\centering
\begin{center}
    \includegraphics[width=\linewidth]{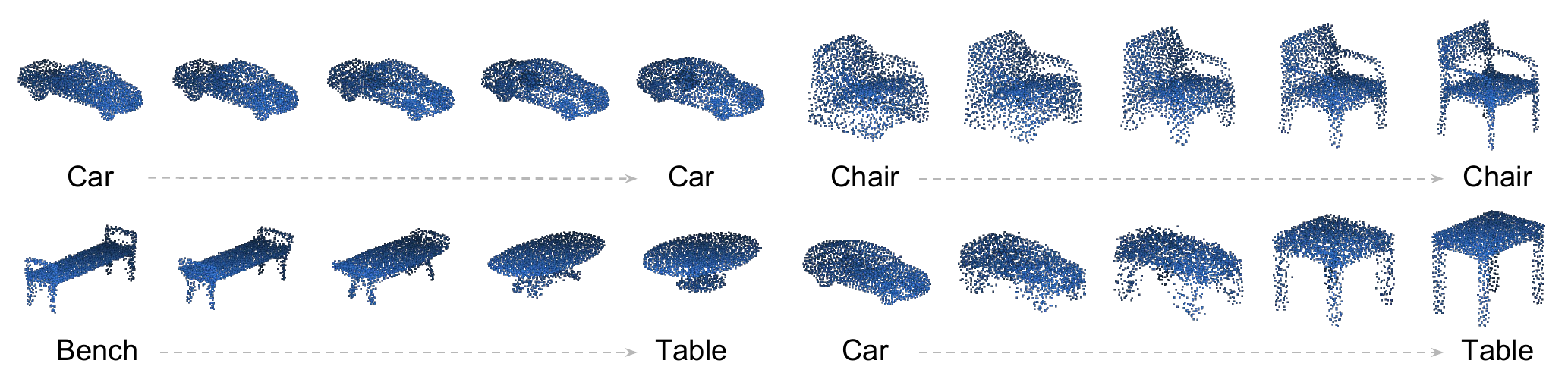}
\end{center}
\caption{Auto-Encoder Interpolations}
\label{fig:ae_interpolations}
\end{figure*}

We analyze the quality of the learnt latent space of the auto-encoder by manipulating the latent vector $z$, and visually observing the generated reconstructions. Fig.\ref{fig:ae_interpolations} shows the resulting reconstructions as we linearly interpolate between two different models in the test set. We find that the interpolations are smooth and the intermediate reconstructions form valid models even in the cross-category setting.

\end{document}